\def\year{2015}
\documentclass[letterpaper]{article}
\usepackage{aaai}
\usepackage{times}
\usepackage{helvet}
\usepackage{courier}
\usepackage{amsfonts,amssymb}
\usepackage{amsmath}
\usepackage{graphicx}
\usepackage{algorithm}           
\usepackage{algorithmic} 
\usepackage{array}
\usepackage{multirow}
\usepackage{url}
\usepackage{latexsym}

\title{TransA: An Adaptive Approach for Knowledge Graph Embedding}
\author{Han Xiao$^1$, Minlie Huang$^1$, Hao Yu$^1$, Xiaoyan Zhu$^1$ \\
	$^1$Department of Computer Science and Technology, State Key Lab on Intelligent Technology and Systems, \\
	National Lab for Information Science and Technology, Tsinghua University, Beijing, China
	}
\date{}

\begin{document}
\maketitle

\begin{abstract}
Knowledge representation is a major topic in AI, and many studies attempt to represent entities and relations of knowledge base in a continuous vector space. Among these attempts, translation-based methods build entity and relation vectors by minimizing the translation loss from a head entity to a tail one. In spite of the success of these methods, translation-based methods also suffer from the oversimplified loss metric, and are not competitive enough to model various and complex entities/relations in knowledge bases. To address this issue, we propose \textbf{TransA}, an adaptive metric approach for embedding, utilizing the metric learning ideas to provide a more flexible embedding method. Experiments are conducted on the benchmark datasets and our proposed method makes significant and consistent improvements over the state-of-the-art baselines.

\end{abstract}

\section{Introduction}
Knowledge graphs such as Wordnet \cite{miller1995wordnet} and Freebase \cite{bollacker2008freebase} play an important role in AI researches and applications. Recent researches such as query expansion prefer involving knowledge graphs \cite{bao2014knowledge} while some industrial applications such as question answering robots are also powered by knowledge graphs \cite{fader2014open}. However, knowledge graphs are symbolic and logical, where numerical machine learning methods could hardly be applied. This disadvantage is one of the most important challenges for the usage of knowledge graph. To provide a general paradigm to support computing
on knowledge graph, various knowledge graph embedding methods have been proposed, such as TransE \cite{bordes2013translating}, TransH \cite{wang2014knowledge} and TransR \cite{lin2015learning}.
 

\begin{figure}
	\centering
	\includegraphics[width=1.0\linewidth]{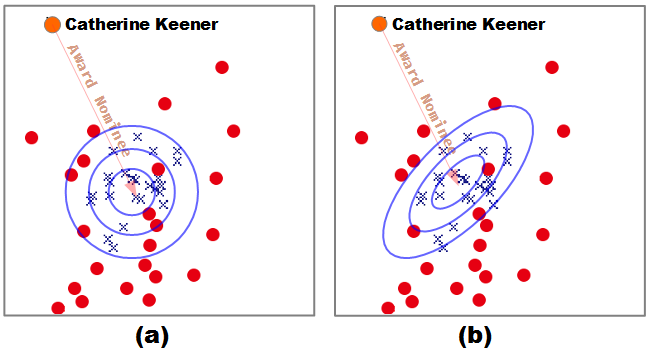}
	\caption{Visualization of TransE embedding vectors for Freebase with PCA dimension reduction. The navy crosses are the matched tail entities for an actor's award nominee, while the red circles are the unmatched ones. TransE applies Euclidean metric and spherical equipotential surfaces, so it must make seven mistakes as (a) shows. Whilst TransA takes advantage of adaptive Mahalanobis metric and elliptical equipotential surfaces in (b), four mistakes are avoided.}
	\label{fig:fig_1}
\end{figure}

Embedding is a novel approach to address the representation and reasoning problem for knowledge graph. It transforms entities and relations into continuous vector spaces, where knowledge graph completion and knowledge classification can be done. Most commonly, knowledge graph is composed by triples $(h,r,t)$ where a head entity $h$, a relation $r$ and a tail entity $t$ are presented. Among all the proposed embedding approaches, geometry-based methods are an important branch, yielding the state-of-the-art predictive performance. More specifically, geometry-based embedding methods represent an entity or a relation as $k$-dimensional vector, then define a score function $f_r(h,t)$ to measure the plausibility of a triple $(h,r,t)$. Such approaches almost follow the same geometric principle $\mathbf{h+r} \approx \mathbf{t}$ and apply the same loss metric $||\mathbf{h+r-t}||_2^2$ but differ in the relation space where a head entity $\mathbf{h}$ connects to a tail entity $\mathbf{t}$. 

However, the loss metric in translation-based models is oversimplified. This flaw makes the current embedding methods incompetent to model various and complex entities/relations in knowledge base.

Firstly, due to the inflexibility of loss metric, current translation-based methods apply spherical equipotential hyper-surfaces with different plausibilities,  where more near to the centre, more plausible the triple is. As illustrated in Fig.\ref{fig:fig_1}, spherical equipotential hyper-surfaces are applied in (a), so it is difficult to identify the matched tail entities from the unmatched ones. As a common sense in knowledge graph, complex relations, such as one-to-many, many-to-one and many-to-many relations, always lead to complex embedding topologies. Though complex embedding situation is an urgent challenge, spherical equipotential hyper-surfaces are not flexible enough to characterise the topologies, making current translation-based methods incompetent for this task.

\begin{figure}
	\centering
	\includegraphics[width=1.0\linewidth]{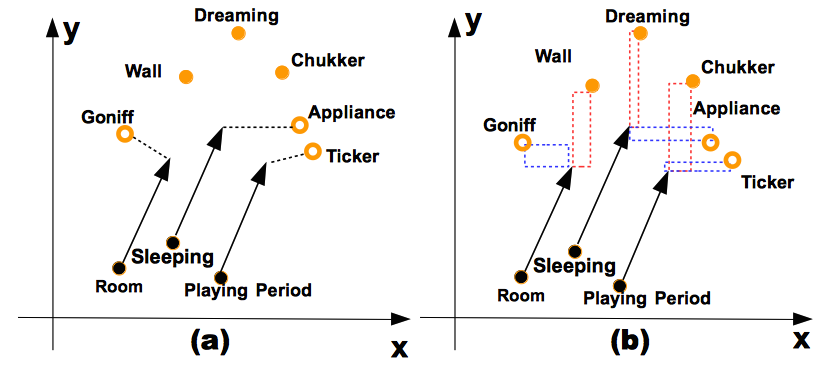}
	\caption{Specific illustration of weighting dimensions. The data are selected from Wordnet. The solid dots are correct matches while the circles are not. The arrows indicate $\mathrm{HasPart}$ relation. (a) The incorrect circles are matched, due to the isotropic Euclidean distance. (b) By weighting embedding dimensions, we up-weighted y-axis component of loss and down-weighted x-axis component of loss, thus the embeddings are refined because the correct ones have smaller loss in x-axis direction.}
	\label{fig:fig_2}
\end{figure}

Secondly, because of the oversimplified loss metric, current translation-based methods treat each dimension identically. This observation leads to a flaw illustrated in Fig.\ref{fig:fig_2}. As each dimension is treated identically in (a)\footnote{The dash lines indicate the x-axis component of the loss $(h_x+r_x-t_x)$ and the y-axis component of the loss $(h_y+r_y-t_y)$.}, the incorrect entities are matched, because they are closer than the correct ones, measured by isotropic Euclidean distance. Therefore, we have a good reason to conjecture that a relation could only be affected by several specific dimensions while the other unrelated dimensions would be noisy. Treating all the dimensions identically involves much noises and degrades the performance.

Motivated by these two issues, in this paper, we propose TransA, an embedding method by utilizing an adaptive and flexible metric. First, TransA applies elliptical surfaces instead of spherical surfaces. By this mean, complex embedding topologies induced by complex relations could be represented better. Then, as analysed in ``Adaptive Metric Approach'', TransA could be treated as weighting transformed feature dimensions. Thus, the noise from unrelated dimensions is suppressed. We demonstrate our ideas in Fig.\ref{fig:fig_1} (b) and Fig.\ref{fig:fig_2} (b). 

To summarize, TransA takes the adaptive metric ideas for better knowledge representation. Our method effectively models various and complex entities/relations in knowledge base, and outperforms all the state-of-the-art baselines with significant improvements in experiments.

The rest of the paper is organized as follows: we survey the related researches and then introduce our approach, along with the theoretical analysis. Next, the experiments are present and at the final part, we summarize our paper.

\section{Related Work}
We classify prior studies into two lines: one is the translation-based embedding methods and the other includes many other embedding methods.

\subsection{Translation-Based Embedding Methods}
All the translation-based methods share a common principle $\mathbf{h+r} \approx \mathbf{t}$, but differ in defining the relation-related space where a head entity $\mathbf{h}$ connects to a tail entity $\mathbf{t}$. This principle indicates that $\mathbf{t}$ should be the nearest neighbour of $\mathbf{(h+r)}$. Hence, the translation-based methods all have the same form of score function that applies Euclidean distance to measure the loss, as follows:
\begin{eqnarray}
	f_r(h,t) = ||\mathbf{h_r + r - t_r}||_2^2 \nonumber
\end{eqnarray}
where $\mathbf{h_r, t_r}$ are the entity embedding vectors projected in the relation-specific space. Note that this branch of methods keeps the state-of-the-art performance.

\begin{itemize}
\item \textbf{TransE} \cite{bordes2013translating} lays the entities in the original space, say $\mathbf{h_r = h}$, $\mathbf{t_r = t}$.
\item \textbf{TransH} \cite{wang2014knowledge} projects the entities into a hyperplane for addressing the issue of complex relation embedding, say $\mathbf{h_r  = h - w_r^\top hw_r}$, $\mathbf{t_r  = t - w_r^\top t w_r}$.
\item \textbf{TransR} \cite{lin2015learning} transforms the entities by the same matrix to also address the issue of complex relation embedding, as: $\mathbf{h_r=M_rh}$, $\mathbf{t_r=M_rt}$.
\end{itemize}

Projecting entities into different hyperplanes or transforming entities by different matrices allow entities to play different roles under different embedding situations. However, as the ``Introduction'' argues, these methods are incompetent to model complex knowledge graphs well and particularly perform unsatisfactorily in various and complex entities/relations situation, because of the oversimplified metric.

\textbf{TransM} \cite{fan2014transition} pre-calculates the distinct weight for each training triple to perform better.

\subsection{Other Embedding Methods}
There are also many other models for knowledge graph embedding.

\textbf{Unstructured Model (UM).} The UM \cite{bordes2012joint} is a simplified version of TransE by setting all the relation vectors to zero $\mathbf{r=0}$. Obviously, relation is not considered in this model.

\textbf{Structured Embedding (SE).} The SE model \cite{bordes2011learning} applies two relation-related matrices, one for head and the other for tail. The score function is defined as $f_r(h,t)=||\mathbf{M_{h,r}h - M_{t,r}t}||_2^2$. According to \cite{socher2013reasoning}, this model cannot capture the relationship among entities and relations.

\textbf{Single Layer Model (SLM).} SLM applies neural network to knowledge graph embedding. The score function is defined as
\begin{eqnarray}
	f_r(h,t)=\mathbf{u_r^\top}g(\mathbf{M_{r,1}h + M_{r,2}t}) \nonumber
\end{eqnarray}
Note that SLM is a special case of NTN when the zero tensors are applied. \cite{collobert2008unified} had proposed a similar method but applied this approach into the language model.

\textbf{Semantic Matching Energy (SME).} The SME model \cite{bordes2012joint} \cite{bordes2014semantic} attempts to capture the correlations between entities and relations by matrix product and Hadamard product. The score functions are defined as follows:
\begin{eqnarray}
	f_r = (\mathbf{M_1h+M_2r+b_1})^\top(\mathbf{M_3t+M_4r+b_2}) \nonumber \\
	f_r = (\mathbf{M_1h \otimes M_2r+b_1})^\top(\mathbf{M_3t \otimes M_4r+b_2}) \nonumber
\end{eqnarray}
where $\mathbf{M_1, M_2, M_3}$ and $\mathbf{M_4}$ are weight matrices, $\otimes$ is the Hadamard product, $\mathbf{b_1}$ and $\mathbf{b_2}$ are bias vectors. In some recent work \cite{bordes2014semantic}, the second form of score function is re-defined with 3-way tensors instead of matrices.

\textbf{Latent Factor Model (LFM).} The LFM \cite{jenatton2012latent} uses the second-order correlations between entities by a quadratic form, defined as $f_r(h,t) = \mathbf{h^\top W_rt}$.

\textbf{Neural Tensor Network (NTN).} The NTN model \cite{socher2013reasoning} defines an expressive score function for graph embedding to joint the SLM and LFM.
\begin{eqnarray}
	f_r(h,t) =\mathbf{u_r^\top}g(\mathbf{h^\top W_{\cdot \cdot r}t + M_{r,1}h + M_{r,2}t + b_r}) \nonumber
\end{eqnarray}
where $\mathbf{u_r}$ is a relation-specific linear layer, $g(\cdot)$ is the $tanh$ function, $\mathbf{W_r} \in \mathbb{R}^{d \times d \times k}$ is a 3-way tensor. However, the high complexity of NTN may degrade its applicability to large-scale knowledge bases.

\textbf{RESCAL.} is a collective matrix factorization model as a common embedding method. \cite{nickel2011three} \cite{nickel2012factorizing}.

\textbf{Semantically Smooth Embedding (SSE).} \cite{guo2015semantically} aims at leveraging the geometric structure of embedding space to make entity representations semantically smooth.

\cite{wang2014knowledge} jointly embeds knowledge and texts. \cite{wang2015knowledge} involves the rules into embedding. \cite{lin2015modeling} considers the paths of knowledge graph into embedding.

\section{Adaptive Metric Approach}
In this section, we would introduce the adaptive metric approach, TransA, and present the theoretical analysis from two perspectives.

\subsection{Adaptive Metric Score Function}
As mentioned in ``Introduction'', all the translation-based methods obey the same principle $\mathbf{h + r \approx t}$, but they differ in the relation-specific spaces where entities are projected into. Thus, such methods share a similar score function.
\begin{eqnarray}
	f_r(h,t) & = & ||\mathbf{h+r-t}||_2^2 \nonumber \\
	& = & (\mathbf{h+r-t})^\top (\mathbf{h+r-t})
\end{eqnarray}
This score function is actually Euclidean metric. The disadvantages of the oversimplified metric have been discussed in ``Introduction''. As a consequence, the proposed TransA replaces inflexible Euclidean distance with adaptive Mahalanobis distance of absolute loss, because Mahalanobis distance is more flexible and more adaptive \cite{wang2014survey}. Thus, our score function is as follows:
\begin{eqnarray}
	f_r(h,t) & = & (|\mathbf{h+r-t}|)^\top \mathbf{W_r} (|\mathbf{h+r-t}|)	
\end{eqnarray}
where $|\mathbf{h+r-t}| \doteq (|h_1+r_1-t_1|,|h_2+r_2-t_2|,...,|h_n+r_n-t_n|)$ and $\mathbf{W_r}$ is a relation-specific symmetric non-negative \textbf{weight matrix} that corresponds to the adaptive metric. Different from the traditional score functions, we take the absolute value, since we want to measure the absolute loss between $\left(\mathbf{h+r}\right)$ and $\mathbf{t}$. Furthermore, we would list two main reasons for the applied absolute operator.

On one hand, the absolute operator makes the score function as a well-defined norm only under the condition that all the entries of $\mathbf{W_r}$ are non-negative. A well-defined norm is necessary for most metric learning scenes \cite{kulis2012metric}, and the non-negative condition could be achieved more easily than PSD, so it generalises the common metric learning algebraic form for better rendering the knowledge topologies. Expanding our score function as an induced norm $N_r(\mathbf{e}) = \sqrt{f_r(h,t)}$ where $\mathbf{e \doteq h+r-t}$. Obviously, $N_r$ is non-negative, identical and absolute homogeneous. Besides with the easy-to-verified inequality $N_r(\mathbf{e_1+ e_2}) = \sqrt{\mathbf{|e_1+e_2|^\top W_r|e_1+e_2|}} \le \sqrt{\mathbf{|e_1|^\top W_r|e_1|}} + \sqrt{\mathbf{|e_2| ^\top W_r|e_2|}} = N_r(\mathbf{e_1}) + N_r(\mathbf{e_2})$, the triangle inequality is hold. Totally, absolute operators make the metric a norm with an easy-to-achieve condition, helping to generalise the representation ability.

On the other hand, in geometry, negative or positive values indicate the downward or upward direction, while in our approach, we do not consider this factor. Let's see an instance as shown in Fig.\ref{fig:fig_2}. For the entity $\mathrm{Goniff}$, the x-axis component of its loss vector is negative, thus enlarging this component would make the overall loss smaller, while this case is supposed to make the overall loss larger. As a result, absolute operator is critical to our approach. For a numerical example without absolute operator, when the embedding dimension is two, weight matrix is [0 1; 1 0] and the loss vector $\left(\mathbf{h+r-t}\right)=(e_1,e_2)$, the overall loss would be $2e_1e_2$. If $e_1 \ge 0$ and $e_2 \le 0$, much absolute larger $e_2$ would reduce the overall loss and this is not desired.

\subsection{Perspective from Equipotential Surfaces}

TransA shares almost the same geometric explanations with other translation-based methods, but they differ in the loss metric. For other translation-based methods, the equipotential hyper-surfaces are spheres as the Euclidean distance defines:
\begin{eqnarray}
||\mathbf{(t-h) - r}||^2_2 = \mathcal{C}
\end{eqnarray}
where $\mathcal{C}$ means the threshold or the equipotential value. However, for TransA, the equipotential hyper-surfaces are elliptical surfaces as the Mahalanobis distance of absolute loss states \cite{kulis2012metric}:
\begin{eqnarray}
\mathbf{|(t-h) - r|^\top W_r |(t-h) - r|} = \mathcal{C}
\end{eqnarray}
Note that the elliptical hyper-surfaces would be distorted a bit as the absolute operator applied, but this makes no difference for analysing the performance of TransA. As we know, different equipotential hyper-surfaces correspond to different thresholds and different thresholds decide whether the triples are correct or not.
Due to the practical situation that our knowledge base is large-scale and very complex, the topologies of embedding cannot be distributed as uniform as spheres, justified by Fig.\ref{fig:fig_1}.  Thus, replacing the spherical equipotential hyper-surfaces with the elliptical ones would enhance the embedding.

As Fig.\ref{fig:fig_1} illustrated, TransA would perform better for one-to-many relations. The metric of TransA is symmetric, so it is reasonable that TransA would also perform better for many-to-one relations. Moreover, a many-to-many relation could be treated as both a many-to-one and a one-to-many relation. Generally, TransA would perform better for all the complex relations.


\subsection{Perspective from Feature Weighting}

TransA could be regarded as weighting transformed features. For weight matrix $\mathbf{W_r}$ that is symmetric, we obtain the equivalent unique form by $LDL~Decomposition$ \cite{golub2012matrix} as follows:
\begin{eqnarray}
	& \mathbf{W_r = L_r^\top D_r L_r} \\
	& f_r = \mathbf{(L_r|h+r-t|)^\top D_r (L_r|h+r-t|)}
\end{eqnarray}
In above equations, $\mathbf{L_r}$ can be viewed as a transformation matrix, which transforms the loss vector $\mathbf{|h+r-t|}$ to another space. Furthermore, $\mathbf{D_r}=diag(w_1, w_2, w_3....)$ is a diagonal matrix and different embedding dimensions are weighted by $w_i$. 

As analysed in ``Introduction'', a relation could only be affected by several specific dimensions while the other dimensions would be noisy. Treating different dimensions identically in current translation-based methods can hardly suppress the noise, consequently working out an unsatisfactory performance. We believe that different dimensions play different roles, particularly when entities are distributed divergently. Unlike existing methods, TransA can automatically learn the weights from the data. This may explain why TransA outperforms TransR although both TransA and TransR transform the entity space with matrices.


\subsection{Connection to Previous Works}
Regarding TransR that rotates and scales the embedding spaces, TransA holds two advantages against it. Firstly, we weight feature dimensions to avoid the noise. Secondly, we loosen the PSD condition for a flexible representation. Regarding TransM that weights feature dimensions using pre-computed coefficients, TransA holds two advantages against it. Firstly, we learn the weights from the data, which makes the score function more adaptive. Secondly, we apply the feature transformation that makes the embedding more effective.

\subsection{Training Algorithm}
To train the model, we use the margin-based ranking error. Taking other constraints into account, the target function can be defined as follows:
\begin{eqnarray}
	\min && \sum_{(h,r,t) \in \Delta} \sum_{(h',r',t') \in \Delta'} [f_r(h,t) + \gamma - f_{r'}(h',t')]_{+} +  \nonumber \\
	&&  \lambda \left(\sum_{r \in R} ||\mathbf{W_r}||_{F}^2 \right) + C \left( \sum_{e \in E}  ||\mathbf{e}||_2^2 + \sum_{r \in R} ||\mathbf{r}||_2^2 \right) \nonumber \\
	s.t. && [\mathbf{W_r}]_{ij} \ge 0
\end{eqnarray}
where $[~\cdot~]_{+} \doteq max(0,~\cdot~)$, $\Delta$ is the set of golden triples and $\Delta'$ is the set of incorrect ones, $\gamma$ is the margin that separates the positive and negative triples. $||~\cdot~||_F$ is the F-norm of matrix. $C$ controls the scaling degree, and $\lambda$ controls the regularization of adaptive weight matrix. The $E$ means the set of entities and the $R$ means the set of relations. At each round of training process, $\mathbf{W_r}$ could be worked out directly by setting the derivation to zero. Then, in order to ensure the non-negative condition of $\mathbf{W_r}$, we set all the negative entries of $\mathbf{W_r}$ to zero. 
\begin{eqnarray}
\mathbf{W_r} = && - \sum_{(h,r,t) \in \Delta} \left( \mathbf{|h+r-t|} \mathbf{|h+r-t|}^\top \right)  \\
+ && \sum_{(h',r',t') \in \Delta'}  \left( \mathbf{|h'+r'-t'|} \mathbf{|h'+r'-t'|}^\top \right) \nonumber
\end{eqnarray}

As to the complexity of our model, the weight matrix is completely calculated by the existing embedding vectors, which means TransA almost has the same free parameter number as TransE. As to the efficiency of our model, the weight matrix has a closed solution, which speeds up the training process to a large extent.

\section{Experiments}
We evaluate the proposed model on two benchmark tasks: link prediction and triples classification. Experiments are conducted on four public datasets that are the subsets of Wordnet and Freebase. The statistics of these datasets are listed in Tab.\ref{tab1}. 

ATPE is short for ``Averaged Triple number Per Entity''. This quantity measures the diversity and complexity of datasets. Commonly, more triples lead to more complex structures of knowledge graph. To express the more complex structures, entities would be distributed variously and complexly. Overall, embedding methods produce less satisfactory results in the datasets with higher ATPE, because a large ATPE means a various and complex entities/relations embedding situation.

\begin{table}
	\centering
	\caption{Statistics of datasets}
	\label{tab1}
	\small
	\renewcommand\arraystretch{1.2}
	\begin{tabular}{|m{0.15\linewidth}<{\centering}|m{0.14\linewidth}<{\centering}|m{0.14\linewidth}<{\centering}|m{0.14\linewidth}<{\centering}|m{0.14\linewidth}<{\centering}|}
		\hline \textbf{Data} & \textbf{WN18} & \textbf{FB15K} & \textbf{WN11} & \textbf{FB13} \\
		\hline
		\hline \#Rel & 18 & 1,345 & 11 & 13 \\
		\hline \#Ent & 40,943 & 14,951 & 38,696 & 75,043 \\
		\hline \#Train & 141,442 & 483,142 & 112,581 & 316,232\\
		\hline \#Valid & 5,000 & 50,000 & 2,609 & 5,908\\
		\hline \#Test & 5,000 & 59,071 & 10,544 & 23,733\\
		\hline
		\hline ATPE \protect\footnotemark & 3.70 & \textbf{39.61} & 3.25 & 4.61 \\
		\hline
	\end{tabular}
\end{table}

\subsection{Link Prediction}

\begin{table*}
	\centering
	\caption{Evaluation results on link prediction}
	\label{tab2}
	\renewcommand\arraystretch{1.1}
	\begin{tabular}{|*{9}{c|}}
		\hline \textbf{Datasets} & \multicolumn{4}{c|}{\textbf{WN18}}  &  \multicolumn{4}{c|}{\textbf{FB15K}}  \\
		\hline
		\hline
		\multirow{2}*{Metric} & \multicolumn{2}{c|}{Mean Rank} & \multicolumn{2}{c|}{HITS@10(\%)} & \multicolumn{2}{c|}{Mean Rank} & \multicolumn{2}{c|}{HITS@10(\%)} \\
		\cline{2-9} & Raw & Filter & Raw & Filter & Raw & Filter & Raw & Filter \\
		\hline
		SE　\cite{bordes2011learning} & 1,011 & 985 & 68.5 & 80.5 & 273 & 162 & 28.8 & 39.8  \\
		SME \cite{bordes2012joint} & 545 & 533 & 65.1 & 74.1 & 274 & 154 & 30.7 & 40.8  \\
		LFM \cite{jenatton2012latent} & 469 & 456 & 71.4 & 81.6 & 283 & 164 & 26.0 & 33.1  \\
		TransE \cite{bordes2013translating} & 263 & 251 & 75.4 & 89.2 & 243 & 125 & 34.9 & 47.1 \\
		TransH \cite{wang2014knowledge} & 401 & 388 & 73.0 & 82.3 & 212 & 87 & 45.7 & 64.4 \\
		TransR \cite{lin2015learning} & \textbf{238} & \textbf{225} & 79.8 & 92.0 & 198 & 77& 48.2 & 68.7 \\
		\hline
		\hline Adaptive Metric (PSD) & 289 & 278 & 77.6 & 89.6 & 172 & 88 & 52.4 & 74.2 \\
		\hline \textbf{TransA} & 405 & 392 & \textbf{82.3} & \textbf{94.3} & \textbf{155} & \textbf{74} & \textbf{56.1} & \textbf{80.4} \\
		\hline
	\end{tabular}
\end{table*}

Link prediction aims to predict a missing entity given the other entity and the relation. In this task, we predict $t$ given $(h, r, *)$, or predict $h$ given $(*, r, t)$. The WN18 and FB15K datasets are the benchmark datasets for this task.

\textbf{Evaluation Protocol.} We follow the same protocol as used in TransE \cite{bordes2013translating}, TransH \cite{wang2014knowledge} and TransR \cite{lin2015learning}. For each testing triple $(h,r,t)$, we replace the tail $t$ by every entity $e$ in the knowledge graph and calculate a dissimilarity score with the score function $f_r(h,e)$ for the corrupted triple $(h,r,e)$. Ranking these scores in ascending order, we then get the rank of the original correct triple. There are two metrics for evaluation: the averaged rank (Mean Rank) and the proportion of testing triples, whose ranks are not larger than 10 (HITS@10). This is called ``Raw'' setting. When we filter out the corrupted triples that exist in all the training, validation and test datasets, this is the``Filter'' setting. If a corrupted triple exists in the knowledge graph, ranking it before the original triple is acceptable. To eliminate this issue, the ``Filter'' setting is more preferred. In both settings, a lower Mean Rank or a higher HITS@10 is better.

\textbf{Implementation.} As the datasets are the same, we directly copy the experimental results of several baselines from the literature, as in \cite{bordes2013translating}, \cite{wang2014knowledge} and \cite{lin2015learning}. We have tried several settings on the validation dataset to get the best configuration for both Adaptive Metric (PSD) and TransA. Under the ``bern.'' sampling strategy, the optimal configurations are: learning rate $\alpha = 0.001$,  embedding dimension $k=50$, $\gamma = 2.0$, $C=0.2$ on WN18; $\alpha = 0.002$, $k=200$,  $\gamma = 3.2$, and  $C=0.2$ on FB15K. 

\begin{table*}
	\centering
	\caption{Evaluation results on FB15K by mapping properties of relations(\%)}
	\label{tab3}
	\renewcommand\arraystretch{1.1}
	\begin{tabular}{|*{9}{c|}}
		\hline \textbf{Tasks} & \multicolumn{4}{c|}{\textbf{Predicting Head(HITS@10)}}  &  \multicolumn{4}{c|}{\textbf{Predicting Tail(HITS@10)}}  \\
		\hline
		\hline
		Relation Category & 1-1 & 1-N & N-1 & N-N & 1-1 & 1-N & N-1 & N-N \\
		\hline
		SE　\cite{bordes2011learning} & 35.6 & 62.6 &17.2 & 37.5& 34.9 & 14.6 & 68.3 & 41.3  \\
		SME \cite{bordes2012joint} & 35.1 & 53.7& 19.0 & 40.3 & 32.7& 14.9 & 61.6 & 43.3 \\
		TransE \cite{bordes2013translating} & 43.7 &65.7 & 18.2 & 47.2 & 43.7 & 19.7 & 66.7 & 50.0 \\
		TransH \cite{wang2014knowledge} & 66.8 & 87.6 & 28.7 & 64.5 & 65.5 & 39.8 & 83.3 & 67.2 \\
		TransR \cite{lin2015learning} & 78.8 & 89.2 & 34.1 & 69.2 & 79.2 & 37.4 & 90.4 & 72.1 \\
		\hline
		\hline \textbf{TransA}& \textbf{86.8} & \textbf{95.4} & \textbf{42.7} & \textbf{77.8} & \textbf{86.7} & \textbf{54.3} & \textbf{94.4} & \textbf{80.6} \\
		\hline
	\end{tabular}
\end{table*}

\footnotetext{ATPE:Averaged Triple number Per Entity. Triples are summed up from all the \#Train, \#Valid and \#Test.}

\textbf{Results.} Evaluation results on WN18 and FB15K are reported in Tab.\ref{tab2} and Tab.\ref{tab3}, respectively. We can conclude that:
\begin{enumerate}
	\item TransA outperforms all the baselines significantly and consistently. This result justifies the effectiveness of TransA. 
	\item FB15K is a very various and complex entities/relations embedding situation, because its ATPE is absolutely highest among all the datasets. However, TransA performs better than other baselines on this dataset, indicating that TransA performs better in various and complex entities/relations embedding situation. WN18 may be less complex than FB15K because of a smaller ATPE. Compared to TransE, the relative improvement of TransA on WN18 is 5.7\% while that on FB15K is 95.2\%. This comparison shows TransA has more advantages in the various and complex embedding environment.
	\item TransA promotes the performance for 1-1 relations, which means TransA generally promotes the performance on simple relations. TransA also promotes the performance for 1-N, N-1, N-N relations\footnote{Mapping properties of relations follow the same rules in \cite{bordes2013translating}.}, which demonstrates TransA works better for complex relation embedding.
	\item Compared to TransR, better performance of TransA means the feature weighting and the generalised metric form leaded by absolute operators, have significant benefits, as analysed.
	\item Compared to Adaptive Metric (PSD)  which applies the score function $f_r(h,t) = \mathbf{(h+r-t)^\top W_r(h+r-t)}$ and constrains $\mathbf{W_r}$ as PSD, TransA is more competent, because our score function with non-negative matrix condition and absolute operator produces a more flexible representation than that with PSD matrix condition does, as analysed in ``Adaptive Metric Approach''.
	\item TransA performs bad in Mean Rank on WN18 dataset. Digging into the detailed situation, we discover there are 27 testing triples (0.54\% of the testing set) whose ranks are more than 30,000, and these few cases would make about 162 mean rank loss. The tail or head entity of all these triples have never been co-occurring with the corresponding relation in the training set. It is the insufficient training data that leads to the over-distorted weight matrix and the over-distorted weight matrix is responsible for the bad Mean Rank.
\end{enumerate}

\subsection{Triples Classification}
Triples classification is a classical task in knowledge base embedding, which aims at predicting whether a given triple $(h,r,t)$ is correct or not. Our evaluation protocol is the same as prior studies. Besides, WN11 and FB13 are the benchmark datasets for this task. Evaluation of classification needs negative labels. The datasets have already been built with negative triples, where each correct triple is corrupted to get one negative triple.

\textbf{Evaluation Protocol.} The decision rule is as follows: for a triple $(h,r,t)$, if $f_r(h,t)$ is below a threshold $\sigma_r$, then positive; otherwise negative. The thresholds $\{\sigma_r\}$ are determined on the validation dataset. The final accuracy is based on how many triples are classified correctly.

\textbf{Implementation.} As all methods use the same datasets,  we directly copy the results of different methods from the literature. We have tried several settings on the validation dataset to get the best configuration for both Adaptive Metric (PSD) and TransA. The optimal configurations are: ``bern'' sampling,  $\alpha= 0.02$, $k=50$, $\gamma=10.0$, $C=0.2$ on WN11, and ``bern'' sampling, $\alpha=0.002$,  $k=200$, $\gamma=3.0$, $C=0.00002$ on FB13. 

\begin{table}
	\centering
	\caption{Triples classification: accuracies(\%) for different embedding methods}
	\label{tab4}
	\renewcommand\arraystretch{1.1}
	\begin{tabular}{|c|c|c|c|}
		\hline \textbf{Methods} & \textbf{WN11} & \textbf{FB13} & \textbf{Avg.} \\
		\hline
		\hline 
		LFM  & 73.8 & 84.3 & 79.0\\
		NTN  & 70.4 & 87.1 & 78.8 \\
		TransE & 75.9 & 81.5 & 78.7 \\
		TransH & 78.8 & 83.3 & 81.1 \\
		TransR & \textbf{85.9} & 82.5 & 84.2\\
		\hline
		\hline Adaptive Metric (PSD)  & 81.4 & 87.1 & 84.3  \\
		\hline \textbf{TransA}& 83.2 & \textbf{87.3} & \textbf{85.3} \\
		\hline
	\end{tabular}
\end{table}

\begin{figure}
	\centering
	\includegraphics[width=1.0\linewidth]{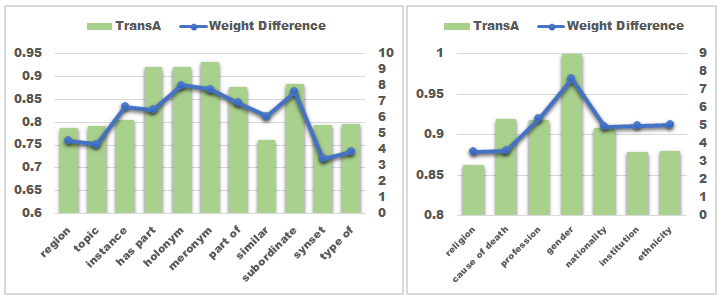}
	\caption{Triples classification accuracies for each relation on WN11(left) and FB13(right). The ``weight difference'' is worked out by the scaled difference between maximal and median weight.}
	\label{fig:fig_3}
\end{figure}

\textbf{Results.} Accuracies are reported in Tab.\ref{tab4} and Fig.\ref{fig:fig_3}. According to ``Adaptive Metric Approach'' section, we could work out the weights by $LDL~Decomposition$ for each relation. Because the minimal weight is too small to make a significant analysis, we choose the median one to represent relative small weight. Thus, ``Weight Difference'' is calculated by
$ \left( \frac{MaximalWeight- MedianWeight}{MedianWeight} \right) $.
Bigger the weight difference is, more significant effect, the feature weighting makes. Notably, scaling by the median weight makes the weight differences comparable to each other. We observe that: 
\begin{enumerate}
\item Overall, TransA yields the best average accuracy, illustrating the effectiveness of TransA.
\item Accuracies vary with the weight difference, meaning the feature weighting benefits the accuracies. This proves the theoretical analysis and the effectiveness of TransA.
\item Compared to Adaptive Metric (PSD) , TransA performs better, because our score function with non-negative matrix condition and absolute operator leads to a more flexible representation than that with PSD matrix condition does.
\end{enumerate}


\section{Conclusion}
In this paper, we propose TransA, a translation-based knowledge graph embedding method with an adaptive and flexible metric. TransA applies elliptical equipotential hyper-surfaces to characterise the embedding topologies and weights several specific feature dimensions for a relation to avoid much noise. Thus, our adaptive metric approach could effectively model various and complex entities/relations in knowledge base. Experiments are conducted with two benchmark tasks and the results show TransA achieves consistent and significant improvements over the current state-of-the-art baselines. 
To reproduce our results, our codes and data will be published in github.



\bibliographystyle{aaai}
\bibliography{aaai}

\begin{thebibliography}{}

\bibitem[\protect\citeauthoryear{Bao \bgroup et al\mbox.\egroup
  }{2014}]{bao2014knowledge}
Bao, J.; Duan, N.; Zhou, M.; and Zhao, T.
\newblock 2014.
\newblock Knowledge-based question answering as machine translation.
\newblock {\em Cell} 2:6.

\bibitem[\protect\citeauthoryear{Bollacker \bgroup et al\mbox.\egroup
  }{2008}]{bollacker2008freebase}
Bollacker, K.; Evans, C.; Paritosh, P.; Sturge, T.; and Taylor, J.
\newblock 2008.
\newblock Freebase: a collaboratively created graph database for structuring
  human knowledge.
\newblock In {\em Proceedings of the 2008 ACM SIGMOD international conference
  on Management of data},  1247--1250.
\newblock ACM.

\bibitem[\protect\citeauthoryear{Bordes \bgroup et al\mbox.\egroup
  }{2011}]{bordes2011learning}
Bordes, A.; Weston, J.; Collobert, R.; Bengio, Y.; et~al.
\newblock 2011.
\newblock Learning structured embeddings of knowledge bases.
\newblock In {\em Proceedings of the Twenty-fifth AAAI Conference on Artificial
  Intelligence}.

\bibitem[\protect\citeauthoryear{Bordes \bgroup et al\mbox.\egroup
  }{2012}]{bordes2012joint}
Bordes, A.; Glorot, X.; Weston, J.; and Bengio, Y.
\newblock 2012.
\newblock Joint learning of words and meaning representations for open-text
  semantic parsing.
\newblock In {\em International Conference on Artificial Intelligence and
  Statistics},  127--135.

\bibitem[\protect\citeauthoryear{Bordes \bgroup et al\mbox.\egroup
  }{2013}]{bordes2013translating}
Bordes, A.; Usunier, N.; Garcia-Duran, A.; Weston, J.; and Yakhnenko, O.
\newblock 2013.
\newblock Translating embeddings for modeling multi-relational data.
\newblock In {\em Advances in Neural Information Processing Systems},
  2787--2795.

\bibitem[\protect\citeauthoryear{Bordes \bgroup et al\mbox.\egroup
  }{2014}]{bordes2014semantic}
Bordes, A.; Glorot, X.; Weston, J.; and Bengio, Y.
\newblock 2014.
\newblock A semantic matching energy function for learning with
  multi-relational data.
\newblock {\em Machine Learning} 94(2):233--259.

\bibitem[\protect\citeauthoryear{Collobert and
  Weston}{2008}]{collobert2008unified}
Collobert, R., and Weston, J.
\newblock 2008.
\newblock A unified architecture for natural language processing: Deep neural
  networks with multitask learning.
\newblock In {\em Proceedings of the 25th international conference on Machine
  learning},  160--167.
\newblock ACM.

\bibitem[\protect\citeauthoryear{Fader, Zettlemoyer, and
  Etzioni}{2014}]{fader2014open}
Fader, A.; Zettlemoyer, L.; and Etzioni, O.
\newblock 2014.
\newblock Open question answering over curated and extracted knowledge bases.
\newblock In {\em Proceedings of the 20th ACM SIGKDD international conference
  on Knowledge discovery and data mining},  1156--1165.
\newblock ACM.

\bibitem[\protect\citeauthoryear{Fan \bgroup et al\mbox.\egroup
  }{2014}]{fan2014transition}
Fan, M.; Zhou, Q.; Chang, E.; and Zheng, T.~F.
\newblock 2014.
\newblock Transition-based knowledge graph embedding with relational mapping
  properties.
\newblock In {\em Proceedings of the 28th Pacific Asia Conference on Language,
  Information, and Computation},  328--337.

\bibitem[\protect\citeauthoryear{Golub and Van~Loan}{2012}]{golub2012matrix}
Golub, G.~H., and Van~Loan, C.~F.
\newblock 2012.
\newblock {\em Matrix computations}, volume~3.
\newblock JHU Press.

\bibitem[\protect\citeauthoryear{Guo \bgroup et al\mbox.\egroup
  }{2015}]{guo2015semantically}
Guo, S.; Wang, Q.; Wang, B.; Wang, L.; and Guo, L.
\newblock 2015.
\newblock Semantically smooth knowledge graph embedding.
\newblock In {\em Proceedings of ACL}.

\bibitem[\protect\citeauthoryear{Jenatton \bgroup et al\mbox.\egroup
  }{2012}]{jenatton2012latent}
Jenatton, R.; Roux, N.~L.; Bordes, A.; and Obozinski, G.~R.
\newblock 2012.
\newblock A latent factor model for highly multi-relational data.
\newblock In {\em Advances in Neural Information Processing Systems},
  3167--3175.

\bibitem[\protect\citeauthoryear{Kulis}{2012}]{kulis2012metric}
Kulis, B.
\newblock 2012.
\newblock Metric learning: A survey.
\newblock {\em Foundations \& Trends in Machine Learning} 5(4):287--364.

\bibitem[\protect\citeauthoryear{Lin \bgroup et al\mbox.\egroup
  }{2015}]{lin2015learning}
Lin, Y.; Liu, Z.; Sun, M.; Liu, Y.; and Zhu, X.
\newblock 2015.
\newblock Learning entity and relation embeddings for knowledge graph
  completion.
\newblock In {\em Proceedings of the Twenty-Ninth AAAI Conference on Artificial
  Intelligence}.

\bibitem[\protect\citeauthoryear{Lin, Liu, and Sun}{2015}]{lin2015modeling}
Lin, Y.; Liu, Z.; and Sun, M.
\newblock 2015.
\newblock Modeling relation paths for representation learning of knowledge
  bases.
\newblock {\em Proceedings of the 2015 Conference on Empirical Methods in
  Natural Language Processing (EMNLP). Association for Computational
  Linguistics}.

\bibitem[\protect\citeauthoryear{Miller}{1995}]{miller1995wordnet}
Miller, G.~A.
\newblock 1995.
\newblock Wordnet: a lexical database for english.
\newblock {\em Communications of the ACM} 38(11):39--41.

\bibitem[\protect\citeauthoryear{Nickel, Tresp, and
  Kriegel}{2011}]{nickel2011three}
Nickel, M.; Tresp, V.; and Kriegel, H.-P.
\newblock 2011.
\newblock A three-way model for collective learning on multi-relational data.
\newblock In {\em Proceedings of the 28th international conference on machine
  learning (ICML-11)},  809--816.

\bibitem[\protect\citeauthoryear{Nickel, Tresp, and
  Kriegel}{2012}]{nickel2012factorizing}
Nickel, M.; Tresp, V.; and Kriegel, H.-P.
\newblock 2012.
\newblock Factorizing yago: scalable machine learning for linked data.
\newblock In {\em Proceedings of the 21st international conference on World
  Wide Web},  271--280.
\newblock ACM.

\bibitem[\protect\citeauthoryear{Socher \bgroup et al\mbox.\egroup
  }{2013}]{socher2013reasoning}
Socher, R.; Chen, D.; Manning, C.~D.; and Ng, A.
\newblock 2013.
\newblock Reasoning with neural tensor networks for knowledge base completion.
\newblock In {\em Advances in Neural Information Processing Systems},
  926--934.

\bibitem[\protect\citeauthoryear{Wang and Sun}{2014}]{wang2014survey}
Wang, F., and Sun, J.
\newblock 2014.
\newblock Survey on distance metric learning and dimensionality reduction in
  data mining.
\newblock {\em Data Mining and Knowledge Discovery}  1--31.

\bibitem[\protect\citeauthoryear{Wang \bgroup et al\mbox.\egroup
  }{2014}]{wang2014knowledge}
Wang, Z.; Zhang, J.; Feng, J.; and Chen, Z.
\newblock 2014.
\newblock Knowledge graph embedding by translating on hyperplanes.
\newblock In {\em Proceedings of the Twenty-Eighth AAAI Conference on
  Artificial Intelligence},  1112--1119.

\bibitem[\protect\citeauthoryear{Wang, Wang, and Guo}{2015}]{wang2015knowledge}
Wang, Q.; Wang, B.; and Guo, L.
\newblock 2015.
\newblock Knowledge base completion using embeddings and rules.
\newblock In {\em Proceedings of the 24th International Joint Conference on
  Artificial Intelligence}.

\end{thebibliography}
\end{document}